\documentclass[10pt, a4paper]{article}
\usepackage{lrec}
\usepackage{multibib}
\usepackage{graphicx}
\usepackage{tabularx}
\usepackage{textcomp}
\usepackage{soul}
\usepackage{titlesec}
\titleformat{\section}{\normalfont\large\bf\center}{\thesection.}{1em}{}
\titleformat{\subsection}{\normalfont\SmallTitleFont\bf\raggedright}{\thesubsection.}{1em}{}
\titleformat{\subsubsection}{\normalfont\normalsize\bf\raggedright}{\thesubsubsection.}{1em}{}
\renewcommand\thesection{\arabic{section}}
\renewcommand\thesubsection{\thesection.\arabic{subsection}}
\renewcommand\thesubsubsection{\thesubsection.\arabic{subsubsection}}

\usepackage{xcolor}
\usepackage{epstopdf}
\usepackage[utf8]{inputenc}
\usepackage{amsmath}
\usepackage{amssymb}
\usepackage{latexsym}
\usepackage{booktabs}
\usepackage{multirow}
\usepackage{graphicx}
\usepackage{enumitem}
\usepackage{CJKutf8}

\usepackage{xstring}
\usepackage{pbox}
\usepackage{todonotes}
\usepackage{makecell}
\newcommand{\ZH}[1]{\begin{CJK}{UTF8}{gbsn}#1\end{CJK}}
\newcounter{examplecounter}
\newenvironment{example}{\begin{quote}%
    \refstepcounter{examplecounter}%
  \textbf{Example \arabic{examplecounter}}%
  \quad
}{%
\end{quote}%
}

\title{Sentence Level Human Translation Quality Estimation with Attention-based Neural Networks}
\name{Yu Yuan, Serge Sharoff}

\address{Nanjing University of Information Science and Technology; University of Leeds\\
         Nanjing, China; Leeds, United Kingdom\\
         hittle.yuan@gmail.com, s.sharoff@leeds.ac.uk\\}

\abstract{
  This paper explores the use of Deep Learning methods for automatic estimation of quality of human translations.  Automatic estimation can provide useful feedback for translation teaching, examination and quality control.  Conventional methods for solving this task rely on manually engineered features and external knowledge.  This paper presents an end-to-end neural model without feature engineering, incorporating a cross attention mechanism to detect which parts in sentence pairs are most relevant for assessing quality.  Another contribution concerns of prediction of fine-grained scores for measuring different aspects of translation quality. Empirical results on a large human annotated dataset show that the neural model outperforms feature-based methods significantly.  The dataset and the tools are available.
  \\ \newline }

\hypersetup{draft}
\begin{document}

\maketitleabstract
\section{Introduction}

Translation quality can be assessed in many different ways \cite{house15}, for example, in the context of MT it is typically assessed in terms of adequacy and fluency \cite{Koehn:2006}. While human evaluation does provide a good estimate of translation quality, it is time consuming, expensive, subjective and not directly applicable to new translations.

	Automatic translation evaluation can be fast, cheap and consistent. A typical method is to compare the similarity between MT output and references, e.g. BLEU \cite{Papineni:2002}.  On the other hand, more recent reference-free approaches to MT Quality Estimation (MTQE), see \cite{bojar:2018:wmt,barrault:2019:wmt}, use machine learning to predict MT quality from linguistic features from the source sentences and MT outputs. The popularity of MTQE is largely driven by the research in MT development and the necessity of evaluating mass output by various types of MT systems. At the same time, automatic human translation estimation (HTQE) has received much less attention, as this is a much more challenging task.

	However, there is a surging need of automating the evaluation of human translation. This task fits into practical scenarios where human translations are scored by experts for certification, course examination and possibly other applications such as self-evaluation in autonomous learning. Translation proficiency test is often a compulsory module in university language and translation programmes at different levels. Language learners and/or trainee translators need to have their work graded in a formative and/or summative evaluation framework. In particular, during the course of learning to translate, trainee translators can have feedbacks from such automatic evaluation systems that are `always there', without the constraints of the fixed working schedule of course instructors. HTQE (particularly fine-grained HTQE) can help in providing quick feedback so that trainees can carry out in-depth diagnostic analyses on their own. In the language service industry, fast turn-around of quality evaluation is also desirable for quality assurance and control. For translation  or localization service users who do not always possess a working bilingual proficiency, they need to have some computational support on their side to determine the quality of the service they paid for.  Nevertheless,  expert human input may not be immediately available. In a different context, large scale translation certification examinations, such as the ATA certification Exam\footnote{https://www.atanet.org/certification/aboutpractice\_test.php}, ITI professional assessment\footnote{https://www.iti.org.uk/membership/professional-assessment},  CATTI\footnote{http://www.catti.net.cn/} require assessment of many submissions. Using automated evaluation can help in reducing the cost of organizing the examination and mitigate the subjectivity of human evaluation in case an automatic evaluation systems can yield reliable judgement of the  quality of input translations. 

	The reference-free MTQE approaches, nevertheless, do not necessarily work well on the task of predicting quality of human translations, since human translators tend to differ from MT in the kinds of errors they make.  There has been some recent work on HTQE \cite{yuan2016} using rich syntactic and semantic features, which are however language- and resource-dependent. To address these shortcomings, we take a different direction and investigate a neural network model for fine-grained HTQE. In particular we propose a customized attention mechanism in order to capture both local and global bilingual quality information. Experiments show that the proposed method outperforms two feature-based methods with $0.22+$ higher correlation with human judgement, maintaining stable performance across four aspects of translation quality.

\section{Related Work}
	Conventional {\bf feature-based methods} have been used for translation quality estimation, particularly for MT. A number of attempts have been made to use machine learned classifiers and regressors for sentence level MT quality in the series of quality estimation shared tasks, predicting indirect quality indexes, such as post-editing effort \cite{Specia_EAMT:2011}, post-editing  distance \cite{specia2010estimating}, post-editing time \cite{koponen2012post}.
	
	Automatic quality estimation of human translations is a newly emerging topic. \newcite{yuan2016} developed a feature set to predict adequacy and fluency of human translations at the document level, which includes comparison between parsed trees, argument roles, phrase alignments, etc. In contrast, \newcite{zhou2019unsupervised} took an unsupervised approach to approximate and grade human translations into different categories using the bidirectional Word Mover's Distance \cite{kusner2015word}.
        
	There has been recent work using {\bf neural models} to compare a target translation with reference(s) in MT evaluation. For example, \newcite{gupta-orasan-vangenabith:2015:EMNLP} use Tree Long Short Term Memory (Tree-LSTM) based networks for reference-based MT evaluation. They propose a method that is competitive to the current complex feature engineering. \newcite{francisco2015}  implemented neural models aiming to select the better translation from a pair of hypotheses, given the reference translation.
	
	Neural models for MT Quality Evaluation have been also recently tested either as Neural Language models on a mixture of n-grams \cite{paetzold16} or a reference-free MTQE prediction model built on quality vectors obtained from parallel corpora \cite{kim16}.

	Often sentence-level MTQE learn to predict translation quality in a indirect manner by ranking translations from best to worst, while learning the direct assessment which matches human evaluators is a challenging task, requiring extensive feature
engineering and suffering from data sparsity, particularly for sentence-level predictions. Compared with discrete models with manual quality features, neural network models  take low-dimensional dense embeddings as the input, which can be trained from a large-scale dataset, thereby overcoming the issue of sparsity, and capture complex non-local syntactic and semantic information that discrete indicator features can hardly encode.

There has been some research on different ways for integration of LSTMs and CNNs, since the two methods for building the neural networks are somewhat complementary.  \newcite{roussinov20lrec} studied the use of LSTMs (or pre-trained transformers) with convolution filters to predict the lexical relations for pairs of concepts, for example, \textit{Tom Cruise} is \textit{an actor} or \textit{a rat} has \textit{a tail}.  Most similar to our work is the study by \cite{zhou16}, which also used a stacked architecture with LSTM followed by two-dimensional pooling to obtain a fixed-length representation for text classification tasks.  Here we contribute by having a novel stacked siamese architecture applied to a different task, namely HTQE.

Therefore, our contribution is two-fold: we work on  a more challenging task \cite{guzman2017machine} than learning the relative ranking of translations or estimating the similarity between candidate translations and references to simulate the scores produced by professional evaluators; we propose a stacked neural networks for {\bf  fine-grained HTQE} without relying on engineered features and many external resources.

\section {Models}
\begin{figure}[!t]
		\centering
		\includegraphics[width=7cm]{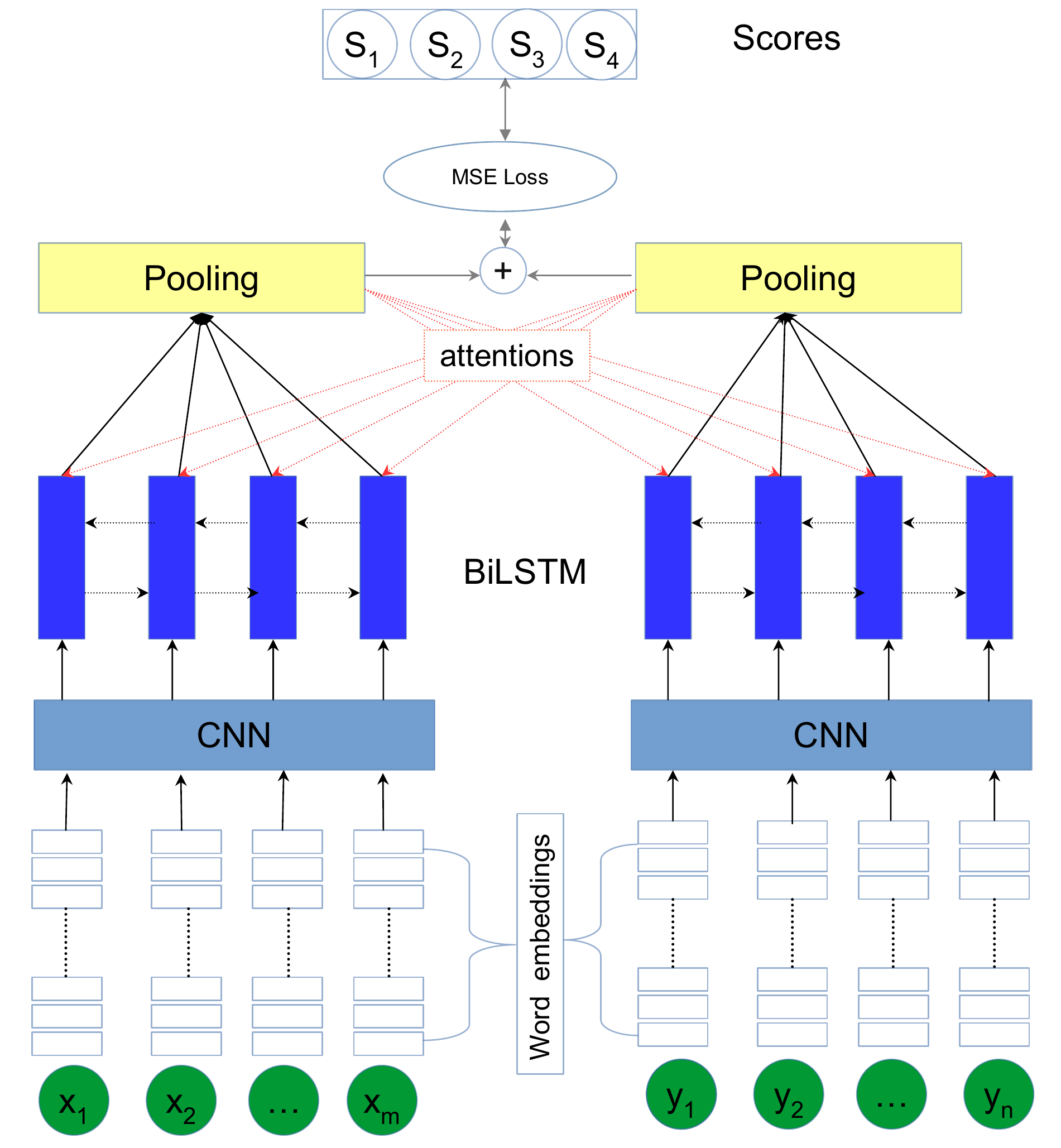}
		\caption{Model Structure}
		\label{fig:model}
	\end{figure}

	Our neural network architecture is shown in Figure \ref{fig:model}. Given a translation pair, the source sentence $x$ and the translated sentence $y$ are encoded into a fixed-sized vector representation through two separate CNN-BiLSTM-Attention stacks. Denoting the final vectors as $\mathbf{x}$ and $\mathbf{y}$ respectively, our model predicts four quality scores (\emph{usefulness}, \emph{terminology}, \emph{idiomatic writing} and \emph{target mechanics} as defined by the ATA, see their definitions below in the Dataset section) using a linear regression on the concatenation of $\mathbf{x}$ and $\mathbf{y}$.
        
	\subsection{Context-aware Word Representation}
	Given  a source sentence $x$ or a translation $y$, which can be represented by $ w_1, w_2, \ldots, w_n $, we first transform the words into vector representations. To this end, we build multiple convolution layers upon standard word embedding layers for context-aware word representation.

	For a convolution layer of width $k$, we apply multiple kernels $\mathbf{H}_i \in \mathbb{R}^{d\times (2k+1)}$ before a non-linearity transformation. Specifically, for a window centred at $i$-th word, the output $\mathbf{f}_{i}$ is given by:
	\begin{equation*}
	\mathbf{f}_{i} = \text{relu}(\langle\mathbf{H}_i, \mathbf{w}_{[i-k:i+k]}\rangle + b_i) \text{,}
	\end{equation*}
	where $\mathbf{w}_{[i-k:i+k]}$ denotes the window size, $ b_i$ is a bias. The word representation is then the concatenation of all convolution layers.
	
	\subsection{Sentence-level Representation}
	To capture global information of a sentence, bidirectional LSTMs \cite{graves2013}  are used on $\mathbf{f}_{i}$. The outputs include a sequence of forward hidden states.
	and a sequence of backward hidden states
	We then concatenate the two sequences into one $h_i = \overrightarrow{h_i}||\overleftarrow{h_i}$ for representing $w_i$. In this way, each annotation $h_i$ contains summarized information about the whole input sentence, but with a strong attention to the details surrounding the $i$-th word.
	
	\subsection{Attention mechanism}
	Different parts in a translation pair do not contribute equally to the semantic adequacy and language fluency of the final output. Attention mechanisms have shown their efficiency in a number of NLP tasks \cite{vaswani17}.  After obtaining the sentence representations centred at different words, we take repeated reading and aligning, using a cross-attention mechanism to detect those bits which are important for quality estimation.

	In particular, we use the weighted average of the source representations to decide which parts of the translated sentence are important for quality estimation and vice versa. Given $h_i$ for each word, the final sentence representation after attention is: 
	\begin{equation}
	\nonumber
	\mathbf{s} = \sum_i^n \alpha_i h_i \text{,}
	\end{equation}
	where $\alpha_i$ is the attention weight for $\mathbf{h}_i$ and it is computed by:
	\begin{equation}
	\nonumber
	\alpha_i = \frac{\exp(f(\mathbf{h}_i, \mathbf{h}))}{\sum_i^n\exp(f(\mathbf{h}_i, \mathbf{h}))}
	\end{equation}
	The score function $f$ is:
	\begin{equation}
	\nonumber
	f(\mathbf{h}, \mathbf{h_i}) = \mathbf{v}^{\text{T}}\tanh(\mathbf{W}_{a1} \overline{\mathbf{h}} + \mathbf{W}_{a2} \mathbf{h}_i) \text{,}
	\end{equation}
	where $\mathbf{v} \in \mathbb{R}^{d_a}$, $\mathbf{W}_{a1}\in \mathbb{R}^{d_a \times 2h} $ and $\mathbf{W}_{a2} \in  \mathbb{R}^{d_a \times 2h}$ are trainable parameters.
	
	\subsection{Training}
	Given a training triple $(x, y, s)$, where $x$ is the source sentence, $y$ is the translated sentence and $s\in \mathbb{R}^k$ is the score vector annotated by human judges from $k$ different aspects, respectively. MSE loss is used for training.
	\begin{equation}
	\nonumber
	\label{eq:loss}
	\ell(x,y,s) = \frac{1}{k}\sum{|\textsc{score}_i(x,y) - s_i|}^2 +\lambda||\Theta||^2
	\end{equation}
	we use Adam \cite{kingma2014adam} to optimize parameters. To avoid over-fitting, dropout is applied with a rate of $0.001$. $\lambda$ is the $l2$ regularization parameter.
\section{Experiments}
	We conduct a set of experiments on the sentence level with a corpus of trainee translation data.

\begin{table}[t]
\begin{center}
\begin{tabular}{lrrrrr}
 & UT & TS & IW & TM & Score\\
\hline
Min. & 2.00 & 2.00 & 3.50 & 1.50 & 11.50\\
1st Quartile & 17.50 & 14.50 & 18.50 & 9.50 & 60.00\\
Median & 23.00 & 18.00 & 20.50 & 11.50 & 71.50\\
Mean & 22.17 & 16.73 & 19.42 & 10.94 & 69.24\\
3rd Quartile & 28.50 & 20.50 & 21.50 & 12.50 & 82.50\\
Max. & 34.50 & 25.00 & 25.00 & 15.00 & 98.50\\
\hline
Krippendorff's $\alpha$ & 0.96 & 0.96 & 0.74 & 0.89 & \\
\end{tabular}
\end{center}
\caption{Description of the dataset}
\label{tab:annot_agreement}
\end{table}

	\subsection{Data Annotation}
	
	 The corpus consists of six source texts selected from the Parallel Corpus of Chinese EFL Learners \cite{wen2008} translated from English into Chinese by learner translators, resulting in 458 translated texts, 3529 Chinese sentences.  We annotated these texts on the sentence level following a percentile scoring scheme according to the American Translators Association (ATA) Certification Programme Rubric for Grading\footnote{\url{http://www.atanet.org/certification/aboutexams\_rubic.pdf}}.  The marks are given for the following four components of translation quality with different weights, i.e. `\emph{usefulness}' (UT) $35$ points,  `\emph{terminology}' (TS) $25$ points, `\emph{idiomatic writing}' (IW) $25$ points and `\emph{target mechanics}' (TM) $15$ points, thus the maximal possible total score is $100$ points.  

     Annotation has been performed by two independent annotators, both teaching translation in China.  The inter-annotator agreement (Krippendorff's $\alpha$) for each of the four components is above $0.74$, see Table~\ref{tab:annot_agreement}.

	\subsection{Setup}

	\begin{table}[!t]
		\centering
		
		\begin{tabular}{ll}
			\hline
			word embedding size   & $d = 200$       \\
			window size           & $k=[1,2,3,4]$   \\
			initial learning rate & $\alpha =0.001$  \\
			dropout rate          & $p = 0.5$       \\
			regularization        & $\lambda =1e-3$ \\
			number of layer       & 1               \\ \hline
		\end{tabular}
		\caption{Hyper-parameter settings}
		\label{tab:hyper}
	\end{table}
	
	We split our data into a training set ($3000$ sentence pairs) and a test set ($529$ sentence pairs). The hyper-parameter settings of our models are listed in Table \ref{tab:hyper}. We use pre-trained word embeddings to initialize the word representations. For English, the pre-trained 200 dimension GloVe vectors \cite{pennington2014} are used. For Chinese, we train a 200 dimension word embeddings on Chinese Wikipedia\footnote{\url{https://dumps.wikimedia.org/zhwiki/}}, using Gensim \cite{rehurek_lrec} with default settings to ensure consistent word segmentation.

        \subsection{Results}

	As traditional in MTQE studies \cite{bojar:2018:wmt}, as well as in HTQE \cite{yuan2016}, we report the correlations of the predicted scores with human judgements using Pearson's $r$ and Spearman's $\rho$ in addition to the mean squared error (MSE).

	Table \ref{tab:experiment_results} presents the results. Note that we experimented with $4$ different window sizes for CNN (See Table \ref {tab:hyper}) and all the neural models reported here use the  window size $2$. We also reproduce the two traditional feature-based methods, i.e. QuEst \cite{specia-etal2013}  with 17 basic features and MoBiL \cite{yuan2016} with nearly 360 features, using XGBoost \cite{Chen:2016} for learning, as it produced better results on this task than other methods \cite{yuyuan2018}. The performance of the neural models without the attention mechanism is also reported in this table.

	\begin{table}[!t] 
		\centering
		\begin{tabular}{llrrr}
			\hline
			{\bf Model}                                  & {\bf Target}            & {\bf$r $}           & {\bf $\rho $}        & {\bf MSE}      \\ \hline
			\multirow{4}{*}{QuEst} & UT  & 0.24   & 0.25    & 51.99 \\
			& TS & 0.08    & 0.09   & 29.26      \\
			& IW & -0.01   & 0.01   & 10.19   \\
			& TM  & -0.01  & 0.01   & 6.07   \\ \hline
			\multirow{4}{*}{MoBiL}  & UT   & 0.18  & 0.20  & 79.23   \\
			& TS & 0.08   & 0.08    & 34.47    \\
			& IW & 0.15   & 0.12    & 16.68        \\
			& TM  & 0.07  & 0.06    & 9.25   \\ \hline
			\multirow{4}{*}{CNN-BiLSTM} & UT   &  0.19   &   0.18  &     64.41     \\
			& TS &  0.21   &  0.21    &  25.65  \\
			& IW &  0.13   &  0.09    &  11.46  \\
			& TM &  0.12   &  0.11    &  5.45  \\ \hline  
			\multirow{4}{*}{CNN-BiLSTM-Att}  & UT  & $0.41$ & $ 0.39$ & $ 40.96 $ \\
			& TS &$ 0.37$  & $0.37$  & $15.58$ \\
			& IW & $0.24$          & $0.21$          & $4.63$     \\
			& TM  & $ 0.30$ & $ 0.28$ & $3.59$ \\ \hline
		\end{tabular}
		
		\caption{Correlation with human judgement}
		\label{tab:experiment_results}
	\end{table}
	
The Wilcoxon signed-ranks test indicates that the neural model with attention has achieved significantly better performance in all aspects of quality estimation (nearly an average of $0.22+$ higher correlation with human judgements) against both MoBiL and QuEst (Z=$-3.02$, p$<0.05$). The model without attention achieves comparable performance to the feature-based models in predicting \emph{Usefulness}, and excels in estimating other types of quality scores. While the feature-based models could not predict \emph{Terminology} (TS) and \emph{Target Mechanics} (TM), the neural models demonstrate superiority in these aspects.  The neural model with attention also produces considerably smaller MSEs in comparison to the two baselines. 

This can be due to the fact that there are relatively fewer effective features concerning target fluency, norms or lexical appropriateness in those baseline models, especially taking into account that the model assesses production of students translating into their native language.  The neural model has leverage some semantic and syntactic information using pre-trained embeddings from very large monolingual corpora.

While hand-crafted features, such as the ratio between the verbs in the source and target segments are designed to capture certain aspects of translation quality for a sentence-translation pair, they are largely de-contextualised. First, the sentence-level representations of the source or target become sparse, because many features such as specific dependence relations do not occur in many sentences.  Second, at the cross-sentence level, the source and target side representations are often equally treated side by side without distinguishing the importance of particular features for interpreting translation errors. In the end, the surface level translation features can be represented in sophisticated ways but often the overall performance of feature-based models is specific to the development set, so the model does not learn generalized parameters to apply them to new translations. 

	In comparison, the proposed neural model intends to address the issue of data sparsity while detecting the semantic, syntactic and even discourse  properties of ST and TT  as prominent features and weighting them globally within and across ST and TT sentences, through the three components of feature extraction via CNN, cross-sentence  association via BiLSTM and global weighting via the attention mechanism. The neural model with attention relies on pre-training knowledge from large monolingual sources that is similar to the bilingual proficiency of a human translator achieved through reading texts in two languages. The CNN, BiLSTM neural networks correspond the bilingual competence and reflect on the translation process to determine what has been important in each instance according to the quality feedback.  It is also advantageous that the QE task can be turned into a multitask-learning for different translation quality aspects, such as Usefulness or Terminology.
	
\subsection{Case Studies}
	\subsubsection{Attention Visualization}
\begin{figure}
		\centering
		\includegraphics[width=8cm]{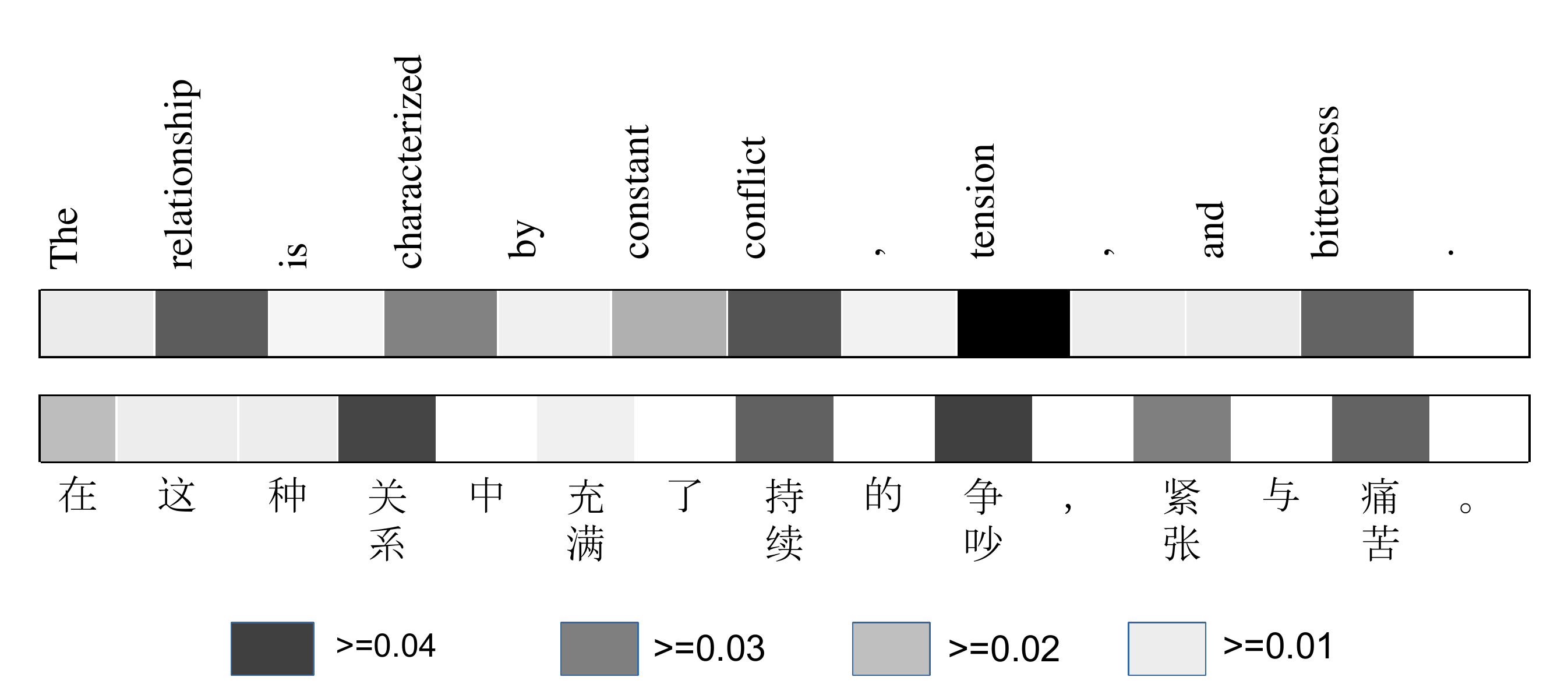}
		\caption{Attention for a Sentence Pair}
		\label{fig:attention}
	\end{figure}

Given the importance of the attention mechanism in our implementation to model HTQE, we visualize a translation pair extracted from the training process, as shown in Figure \ref{fig:attention}. 
	
	The attention mechanism in our approach, as manifested by the plotted weights, does not seek monotonic or predictive alignment as it happens in Neural Machine Translation \cite{luong-pham-manning:2015:EMNLP}. The weights for words in the English source sentence and the Chinese target sentence are not necessarily `aligned' unlike in traditional NMT attention models. This relaxation is advantageous to the task, given that first we have much less data in our quality estimation training set in comparison to NMT parallel corpora.  More importantly, even though aligned segments are indicative of translation quality, they do not contribute equally to the final quality of a translation segment. For example, given a batch of sentences, with all the essential components such as nouns, verbs, adjectives, adverbs, terms, named entities properly aligned to the source sentence, what distinguishes them with respect to translation quality are maybe the trivial details in each translation, e.g. word order, connectives, etc.

        In our experience, content words in both source and target sentences are especially helpful. For instance, the Chinese word \ZH{紧张} (`tension') is weighted less than its correspondence `tension' in English, and neither the English verb `characterize' nor its translation \ZH{充满} (`full of') are selected as important elements by the model. We also notice in this example that the Chinese translation contains \ZH{在} (`in'), which does not exist in the English source, but it is picked up by the attention mechanism. Adding this word to the translation improves its fluency, making the target translation more readable.

	\begin{table*}[!t]
		\centering
		
		{ \begin{tabular}{@{}llllll@{}}
				\toprule
				& {\bf Model} & {\bf UT} & {\bf TS} & {\bf IW} & {\bf TM} \\ \midrule
				\multirow{4}{*}{\parbox{11cm}{Freedom from this constraint is the dream of every transplant surgeon . \\ \ZH{打破 这 种 局限性 的 梦想 就 寄托 在 了 每次 移植 手术 上 了 。}}} & Human         & 6          & 4.5        & 21.5       & 12.5       \\
				& MoBiL         & 17.7       & 13.7       & 16.7       & 9.4        \\
				& QuEst         & 23.3       & 16.4       & 18.0       & 9.2        \\
				& Neural        & {\bf 12.6}      & {\bf 10.5}      & {\bf 20.9}      & {\bf 11.9}      \\ 
				\midrule
				\multirow{6}{*}{\parbox{11cm}{So far attempts to make artificial organs have been disappointing: nature is hard to mimic. hence the renewed interest in trying to use organs from animals\\ \ZH{  到 目前 为止 ， 尝试 模拟 人造 器官 的 结果 让 人 颇 有些 失望 ： 自然 难以 模拟 。 因而 人们 将 更 多 的 目光 投向 动物 的 器官 上 。\\\\}}} & Human & 33.5          & 22.5       & 22.5       & 13       \\[0.6pt]
				& MoBiL         & 21.6       & 17.2       & \textbf{19.5}
                                & 10.5        \\[0.6pt]
				& QuEst         & 22.9       & 18.0       & 19.3       & \textbf{10.8}
                                \\[0.6pt]
				& Neural        & {\bf 26.7}      & {\bf 19.4}      & 16.9      & 10.6      \\ [0.6pt]

				\bottomrule
		\end{tabular}}
		\caption{Human Annotation and Model Predictions}
		\label{tab:examples}
	\end{table*}
	
Therefore, the attention mechanism in the neural architecture is essential as it tries to pinpoint which segments of ST and TT are influential to the final quality judgement. Specifically, by picking a fragment of the ST sentence, the attention mechanism can force the encoding layer (BiLSTM) to understand the importance of this fragment to the final quality when all the available fragments in the TT have been seen, and vice versa, by picking the important fragment of the TT sentence, it forces the encoding layer to understand its importance to the final quality judgement when all the fragments in the ST are seen.  In the end, the equivalent fragments in a ST-TT sentence pair can be weighted differently since the quality estimation process is no longer treated as a sequence-to-sequence learning that the encoder layer reads the source sequence representations and the output layer estimates the conditional probabilities of the target sequence. Instead, the proposed neural network reads ST and TT sequence to predict their joint conditional probability while focusing on which ST or TT representation helps in determining the quality. As shown in Table \ref{tab:experiment_results}, this design significantly boosts the performance of neural model in predicting the four quality labels. In some sense, it is similar to the analytical scoring of human translations when evaluators decompose  a ST-TT pair into several scoring points. However, it is also different  in that in analytical scoring eqaul attention is paid to the equivalents of ST-TT segments. We admit that the present attention design is particularly aimed to highlight segments on both sides and we do know for sure whether it is worth imposing equal weighting between segment pairs. It would be interesting to investigate the influence of different attention strategies on QE in the future. 

\subsubsection{Model Predictions}

	In the upper example of Table \ref{tab:examples}, the neural model with attention predicts the scores for `IW' and `TM' fairly  accurately, which are about the fluency of the translation. As the Chinese translation itself reads rather fluent in terms of language itself and conforms to the Chinese norms, both human annotators and our model assign relatively high and close scores for them for \emph{Idiomatic Writing} and \emph{Target Mechanics}. 
	
	Even though the neural model offers the best prediction for `UT' and `TS', which are about the adequacy of a translation, the differences between the model estimation and human annotation are still significant. A closer examination of the translations reveals that the translation has twisted the meanings of the source sentences due to mistranslations of the English word `surgeon' as \ZH{手术} (`surgery'). In addition, the Chinese word \ZH{寄托} (`place on'), which does not exist in the original, has changed the meaning of the translation. As a consequence, the whole sentence needs to be retranslated, which explains the low score by human annotators for \emph{Usefulness}. Such semantic intricacy requires a model to capture the underlying meaning of sentences, which can impose challenges to manual features. It is the same case with the second example, in which \ZH{更多 的 目光 投向} (`set eyes on') is a non-literal  but valid translation for `renewed interest in'. We suspect that the proposed neural model based on word representations may be biased towards word level adequacy, while significant changes of meaning due to addition, omission and mistranslation to close synonyms could not be detected accurately. For those underscored `good' translations, the same reason applies. In the lower example in Table \ref {tab:examples},  \ZH{结果} (`result'), \ZH{颇} (`rather') and  \ZH{更多 的 目光} (`set eyes on',`derived from renewed interest in') could cause confusion for a model based on word representations. Thus, the neural model has limited validity for adequate scoring of free but still valid translations.

\subsection{Comparison of HT and MT} 
	Another factor closely related to translation quality is the distribution of translation errors both human translations and machine translations contain. The distribution of translation errors in the two modes of translations displays very different patterns. \newcite{vilar2006error}  carried out error analysis on three statistical machine translation engines.  They show that the most common MT errors are missing words, word order and incorrect words as valid across two language directions (English-Spanish and Chinese-English).  In contrast, the most common HT errors are undertranslation (a translation is less specific in comparison to the original), awkward style and syntactic issuesaccg to a statistical corpus-wise comparison of translations errors in HTs and MTs \cite{yuyuan2018}.  To complement the study of translation quality, we show how translation quality variation is embodied in the distribution of translation errors.  For this task we use the adapted DQF-MQM framework \cite{lommel2014multidimensional} to annotate the translations since the framework is explicitly designed for describing both MT and HT quality. The final list of error types used for annotating the data is included below:
	
\begin{itemize}[noitemsep]
    \item {\bf mistranslation} that the target content does not accurately represent the source content.
    \item{\bf omission} that content present in the source is missing from the translation.
    \item{\bf awkward} that a text is written with an awkward style.
    \item{\bf punctuation} that punctuation is misused for the target language.
    \item{\bf undertranslation} that the target text is less specific than the source text.
    \item{\bf unidiomatic} that the content is semantically correct but not as natural as native target texts.
    \item{\bf grammar} that the target text manifests grammatical and/or syntactic fallacies.
    \item{\bf addition} that the target text includes content not present in the source.
    \item{\bf spelling} that the target text has deficient written forms, e.g. spelling error, made-up words.
    \item{\bf terminology} that a domain-specific word is translated into an inappropriate term or a non-term.
    \item{\bf untranslated} that content that should have been translated has been left untranslated.
\end{itemize}
	
To compare the error distribution in MTs and HTs, we translated the six STs of our corpus from English into Chinese using 7 commercial  MT systems and we randomly selected 7 HTs of each source text to form a corpus comparable to MTs.  

Their manual annotation shows that that the most common categories of translation errors are mistranslation, omission, awkward and unidiomatic for both human and machine translations. It is also noteworthy that certain error types, such as grammar and untranslated are more serious in MTs.  The errors are illustrated through the following examples:

\begin{example}[MT-Grammar Error]
    from the top of the mountain , sloping for several acres across folds and valleys were rivers of daffodils in radiant bloom .\\
    \ZH{从 山顶 开始 ， 倾斜 几 英 亩 [awkward] 的 褶皱 [mistranslation] 和 山谷 是 水仙花 盛开 的 水仙花 [grammar]}\\
    gloss: from top of mountain starting , slope several acres folds and valleys are daffodils in blossom daffodils.\\
\end{example}

\begin{example}[HT-Grammar Error]
people already kill pigs both for food and for sport ; killing them to save a human life seems , if anything , easier to justify. however , the science of xenotransplantation is much less straightforward .\\
\ZH{人们 为了 食物 和 运动 的 目的 而 杀 了 很多 猪 。 但是 若 任何 事 都 可以 轻易 地 使之 合理化[mistranslation] ， 人们 杀猪 而 为 自身 的 生存 也 是 合理 合理 的[grammar] 。 况且 ， 异种 器官 移植 的 科学 也 变得 简单 ， 易懂 了[mistranslation] }\\
gloss: people for food and sports  purpose  to kill  many pigs . but  if anything can be easy to be justified ,  people  kill pigs for their existence too is reasonable . and, xenotransplantation  science of  too became easier , more understandable \\
\end{example}

\begin{example}[MT-Omission]
    bees , wasps , ants and termites have intricate societies in which different members are specialized for foraging , defense and reproduction .\\
    \ZH{蜜蜂 、 黄蜂 、 蚂蚁和 白蚁 有 复杂 的 社会 [omission]不同 成员 觅食 是 专用 于 、 国防 和 复制 [mistranslation]。}\\
    gloss: bees , wasps , ants and termites  have complex  societies different members looking for food is specialized  for  , defence  and copy .\\
\end{example}

\begin{example}[HT-Omission]
in Europe and America , herds of pigs are being specially bred and genetically engineered for organ donation . \\
\ZH{在 欧洲 和 美国 为 器官 捐赠 饲养 出 了[mistranslation] 成群 的 受过 特殊 饲养 的 猪[omission] 。}\\
gloss: in Europe and America for organ donation have kept  herds of been specially bred  pigs . \\

\end{example}

The above four examples (2 HTs and 2 MTs) contain 2 instances of omission and 2 instances of grammar errors. In the first example, \ZH{水仙花 盛开 的 水仙花} (`daffodils in blossom daffodils') is ungrammatical because the MT system does not linke the `slope' with `daffodils' and give it a more idiomatic translation \ZH{绵延} (`stretches'), in addition to \ZH{倾斜 几 英 亩} (`slope several acres') that reads very awkward  due to the failure to translate the metaphoric `rivers of daffodils'.
In the fourth example, \ZH{饲养 出 了} (`have kept') mistranslated the present progressive tense `being specially bred', in addition to the \ZH{受过 特殊 饲养 的 猪} (`specially bred pigs') that has omitted the modifier `genetically engineered'. Other two examples contain the similar errors of mistranslation and omission.

\begin{figure}[t]
    \centering
    \includegraphics[width=8cm, height=8cm]{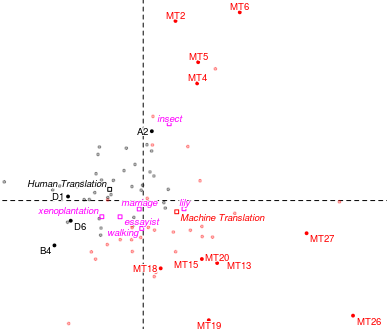}
    \caption{Translations in the first two PCA dimensions}
    \label{fig:contrib}
\end{figure}

We performed the PCA analysis \cite{abdi-williams2010} of the vector of translation error counts, using the varimax\footnote{an orthogonal method to scale the respective eigenvalues by the squared roots so as to obtain the eigenvectors as loadings} rotation method. This helped to identify three underlying dimensions characteristic of the distribution of translations errors in HTs and MTs: language use (first dimension), content inadequacy (second dimension) and lexical misuse (third dimension) from the space of factor loadings of each error types.  Figure \ref{fig:contrib} illustrates the distribution of text topics (in pink) and translation instances (HTs in black and MTs in red) along the first two dimensions. Note that both HTs and MT with contributive importance in term of cosine squared less than $0.5$ are shaded (dark and light black dots are HTs projected on the dimension with smaller cosine squared and so are the dark and light red dots for MTs).   Our data  has shown that the first dimension, i.e. language misuse, characterizes most MTs (MT+ Arabic number  indicates a numbered MT of the 42 MTs ), as top contributive translations to this dimension comprise mainly MTs. In contrast, HTs  (HT + Arabic number indicates a numbered HT sample of the 42 HTs)  centre towards the second dimension, i.e. content inadequacy. These findings suggest that deficiency of HTs in quality may have to do with translators’ inability of delivering the ST content in a sufficient manner. For MTs, these findings imply that language problems, such as grammaticality, naturalness, are typical. These findings echo the findings of  \newcite{vilar2006error}, who also maintain that language issues, such as wrong lexical choice, incorrect form, extra words, style and idiom, are the primary sources of Chinese-English errors. 

	The pattern of HT errors (content inadequacy) implies that HT quality issues arise mainly due to either translators’ decision-making (e.g. undertranslation is a result of translation strategy) or their incapability of switching between two languages (e.g. awkward translations). In contrast, MT errors are more about language misuse, while the subtle difference between ‘good’ and ‘bad’ for human translations are often harder to detect automatically.

	\section{Conclusions}
	This paper presents a neural model for the Human Translation Quality Estimation (HTQE) task, which involves a weighted cross attention mechanism to adaptively detect the relevant parts in the source-target sentence pairs.  Despite having no hand-crafted features, experimental results show that the neural model with attention can outperform conventional feature-based methods as well as a baseline neural model. 
	To our knowledge, we are the first to apply neural networks to reference-free fine-grained HTQE.  Our code and dataset of expert-annotated translations with fine-grained scores for the English-Chinese direction is available under a permissive licence.\footnote{\url{https://github.com/hittle2015/NeuralTQE}}
	
        In the future, we plan expanding this study in two directions. While initial experiments with BERT \cite{devlin18} did not show improvements in the model, we will try truly cross-lingual language models such as XML-R \cite{conneau19unsupervised}, since cross-lingual language models are likely to be more effective in comparison to the current model which uses independent embeddings for each language, while the training set itself is too small to infer links between languages from bilingual data.  Next, we will experiment with the integration of other features into attention, such as alignment information from large parallel corpora, to introduce quality vectors similarly to \cite{kim16}.  Even though the neural architecture outperforms feature-based methods, we can try integrating features which manifest translators' decision-making into the neural network.

\section{Acknowledgements}
This research funded and forms part of the achievements of the Philosophy and Social Sciences Foundation Project of Jiangsu Province (SK20180032) and the general project of Humanities and Social Sciences Foundation of the Ministry of Education (19YJC740114)

\section{References}\label{reference}

\bibliographystyle{lrec}
\bibliography{lrec2020}


\end{document}